\documentclass[final]{cvpr}

\usepackage{times}
\usepackage{epsfig}
\usepackage{graphicx}
\usepackage{amsmath}
\usepackage{amssymb}
\usepackage{booktabs}
\usepackage{multirow}
\usepackage{helvet}
\usepackage{courier}
\usepackage{algpseudocode}
\usepackage{algorithm,tabularx}
\usepackage{diagbox}

\usepackage[pagebackref=true,breaklinks=true,colorlinks,bookmarks=false]{hyperref}
\newcommand{\multiline}[1]{%
  \begin{tabularx}{\dimexpr\linewidth-\ALG@thistlm}[t]{@{}X@{}}
    #1
  \end{tabularx}
}

\newcolumntype{C}[1]{>{\centering\arraybackslash}p{#1}}
\graphicspath{{cambriafig/}}

\def\cvprPaperID{587} 
\def\confYear{CVPR 2021}
\begin{document}

\title{Cycle4Completion: Unpaired Point Cloud Completion using Cycle Transformation with Missing Region Coding}

\author{Xin Wen\textsuperscript{1}, Zhizhong Han\textsuperscript{2}, Yan-Pei Cao\textsuperscript{3}, Pengfei Wan\textsuperscript{3}, Wen Zheng\textsuperscript{3}, Yu-Shen Liu\textsuperscript{1}\thanks{Corresponding author. This work was supported by National Key R\&D Program of China (2020YFF0304100, 2018YFB0505400), the National Natural Science Foundation of China (62072268), and in part by Tsinghua-Kuaishou Institute of Future Media Data.}\\
\textsuperscript{1}School of Software, BNRist, Tsinghua University, Beijing, China\\
\textsuperscript{2}Department of Computer Science, Wayne State University, USA\\
\textsuperscript{3} Y-tech, Kuaishou Technology, Beijing, China\\
{\small x-wen16@mails.tsinghua.edu.cn\hspace{1mm}
h312h@wayne.edu\hspace{1mm}caoyanpei@gmail.com\hspace{1mm}}\\
{\small\{wanpengfei,zhengwen\}@kuaishou.com\hspace{1mm}
liuyushen@tsinghua.edu.cn}
}

\maketitle
\thispagestyle{empty}

\begin{abstract}
  In this paper, we present a novel unpaired point cloud completion network, named Cycle4Completion, to infer the complete geometries from a partial 3D object.
  Previous unpaired completion methods merely focus on the learning of geometric correspondence from incomplete shapes to complete shapes, and ignore the learning in the reverse direction, which makes them suffer from low completion accuracy due to the limited 3D shape understanding ability.
  To address this problem, we propose two simultaneous cycle transformations between the latent spaces of complete shapes and incomplete ones.
  Specifically, the first cycle transforms shapes from incomplete domain to complete domain, and then projects them back to the incomplete domain.
  This process learns the geometric characteristic of complete shapes, and maintains the shape consistency between the complete prediction and the incomplete input.
  Similarly, the inverse cycle transformation starts from complete domain to incomplete domain, and goes back to complete domain to learn the characteristic of incomplete shapes.
  We experimentally show that our model with the learned bidirectional geometry correspondence outperforms state-of-the-art unpaired completion methods. Code will be available at \url{https://github.com/diviswen/Cycle4Completion}.
\end{abstract}

\section{Introduction}
Point clouds, as a popular 3D representation, can be easily produced by 3D scanning devices and depth cameras. However, due to the limitations of the view angles of camera/scanning devices and self-occlusion, raw point clouds are often sparse, noisy and partial, which usually require shape completion before being analyzed in further applications such as shape classification \cite{qi2017pointnet,han20193dviewgraph,9318534,liu2020lrc,liu2019l2g}, retrieval \cite{han2019view,han2019parts,han2019seqviews2seqlabels}, semantic/instance segmentation \cite{liu2019sequence,Wen2020MM}. Although the recent data-driven supervised completion methods \cite{hutaoaaai2020,yuan2018pcn,xie2020grnet,yin2018p2p,huang2020pf,liu2019morphing} have achieved impressive performance, they heavily rely on the paired training data, which consists of incomplete shapes and their corresponding complete ground truth.
In real-world applications, however, such high quality and large-scale paired training dataset is not easy to access, which makes it hard to directly train a supervised completion network.

A promising but challenging solution to this problem is to learn a completion network in an unpaired way, where the common practice is to establish the shape correspondence between the incomplete shapes and complete ones from the unpaired training data without requiring the incomplete and complete correspondence.
The latest work like Pcl2Pcl \cite{chen2019unpaired} introduced an adversarial framework to merge the geometric gap between the complete shape distribution and incomplete one in the latent representation space.
Although many efforts have been made to learn the geometric correspondence from incomplete shapes to complete ones, previous methods ignore the inverse correspondence from complete shapes to incomplete ones, which leads to low completion accuracy due to the limited 3D shape understanding ability.

\begin{figure*}[!t]
  \centering
  \includegraphics[width=0.9\textwidth]{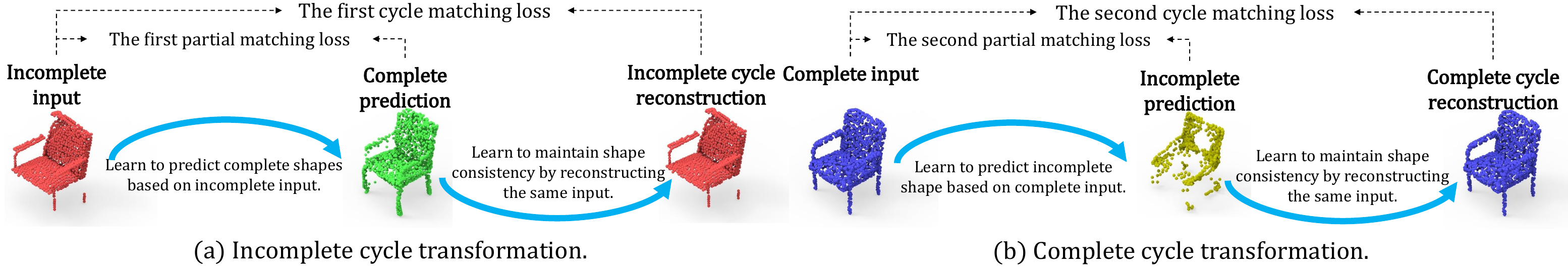}
  \caption{Illustration of cycle transformation, which consists of two inverse cycles, as shown in (a) and (b). The cycle transformation promotes network to understand 3D shapes by learning to generate complete or incomplete shapes from their complementary ones.
  }
  \label{fig:cycle_illustration}
\end{figure*}

\begin{figure}[!t]
  \centering
  \includegraphics[width=0.9\linewidth]{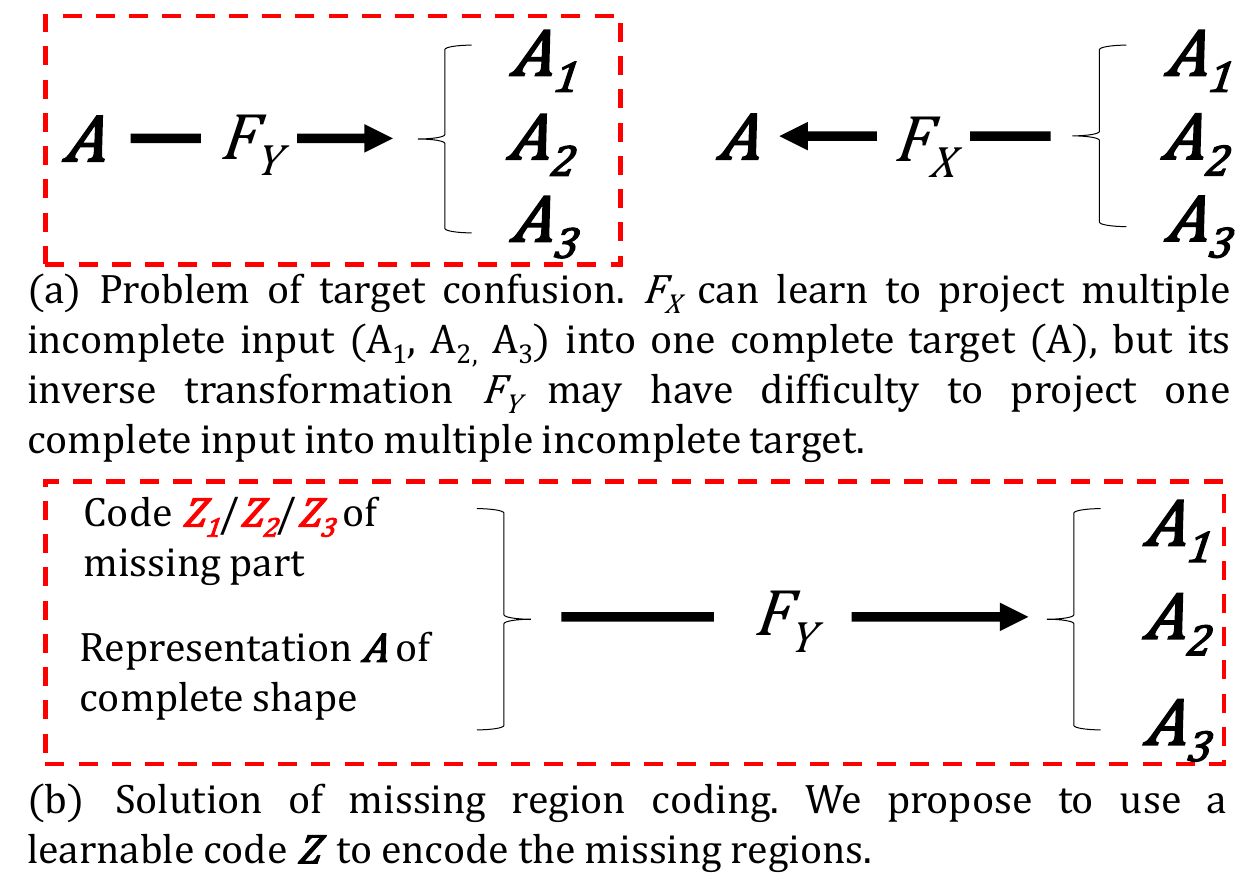}
  \caption{Illustration of target confusion and the solution of missing region coding.
  }
  \label{fig:problem}
\end{figure}

To address this problem, we propose a novel unpaired point cloud completion network, named Cycle4Completion, to establish the geometric correspondence between incomplete and complete shapes in both directions.
We achieve this by designing two \emph{cycle transformations}, i.e. the incomplete cycle transformation (incomplete-cycle) and the complete cycle transformation (complete-cycle), as shown in Figure \ref{fig:cycle_illustration}. The incomplete-cycle in Figure \ref{fig:cycle_illustration}(a) learns the mapping from the incomplete domain to the complete one, which is then projected back to the incomplete domain.
On the other hand, the complete-cycle in Figure \ref{fig:cycle_illustration}(b) provides the completion knowledge on the inverse direction with incomplete input, which can be used to further enhance the incompletion quality for incomplete-cycle.
However, as shown in Figure \ref{fig:problem}(a), directly applying a cycle transformation in the latent space will encounter a new problem which we name it as the \emph{target confusion} problem.
This problem is raised when establishing shape correspondence from multiple incomplete shapes (e.g. $A_1, A_2$ and $A_3$) to one complete shape (e.g. $A$). This is because one of the cycle requires the network to predict the incomplete shape based on the complete input, and the corresponding transformation network $F_Y$ cannot fully map one complete input into multiple different incomplete targets only through a deep neural network.
To solve this problem, we propose the learnable \emph{missing region coding} (MRC) to transform incomplete shapes to complete ones, as shown in Figure \ref{fig:problem}(b). The representations of incomplete shapes can be decomposed into two parts: one is the representation $A$ of their corresponding complete shape, and the other one is the code $Z$ to encode their missing regions. When predicting the complete shapes from the incomplete ones, only the representation $A$ is considered, and when predicting the incomplete shapes from the complete ones, both the representation $A$ and code $Z$ are considered. Thus, the transformation network $F_Y$ will relieve the confusion by learning to project one complete input to several incomplete targets. Instead, the learnable missing region code $Z$ can help the network clarify which incomplete shape is the current target for transformation, and relieve the target confusion problem.
Our main contributions are summarized as follows.
\begin{itemize}
  \item We propose a novel unpaired point cloud completion network, named Cycle4Completion. Compared with previous unpaired completion methods which only consider the single-side correspondence from incomplete shapes to complete ones, Cycle4Completion can enhance the completion performance by establishing the geometric correspondence between complete shapes and incomplete shapes from both directions.
  \item We propose the \emph{cycle transformation} framework in latent space, which is combined with the \emph{partial matching loss} and \emph{cycle matching loss} to establish the bidirectional geometric correspondence between the complete and incomplete shapes, and maintain the shape consistency throughout the whole transformation process.
  \item We propose the \emph{missing region coding} to decompose the incomplete shape representation into a representation of its corresponding complete shape, and a missing region code to encode the missing regions of the incomplete shapes, respectively. This solves the \emph{target confusion} when the network tries to predict multiple incomplete shapes based on a single complete shape.
\end{itemize}

\section{Related Work}
3D shape completion has drawn an increasing attention in recent years \cite{yuan2021survey,wen2020sa,wang2020cascaded}. Previous completion methods can be roughly divided into two categories, i.e. traditional approaches and deep learning based approaches, which we will detail below.

\noindent\textbf{Traditional approaches for 3D shape completion.}
The traditional geometry/statistic based methods \cite{sung2015data,berger2014state,wei2019local,shao2012interactive,martinovic2013bayesian,shen2012structure} exploit the geometric features of surface on the partial input to generate the missing regions of 3D shapes \cite{sung2015data,berger2014state,thanh2016field,wei2019local}, or exploit the large-scale shape database to search for the similar shapes/patches to fill the missing regions of 3D shapes \cite{shao2012interactive,kalogerakis2012probabilistic,martinovic2013bayesian,shen2012structure}. For example, Hu et al.\cite{wei2019local} proposed to exploit both the local smoothness and the non-local self-similarity in point clouds, by defining the smoothing and denoising properties of point clouds and globally searching the similar area for the missing region. On the other hand, the data-driven shape completion methods like Shen et al.\cite{shen2012structure} formulate the completion of 3D shapes as a bottom up part assembling process, where a 3D shape repository is adopted as the reference to recover a variety of high-level complete structures.
In all, these traditional shape completion approaches are mainly based on the hand-crafted rules to describe the characteristics of missing region, and the similarities between the missing region and complete shape. Therefore, the generalization ability of such kind of methods is usually limited. For example, the method proposed by Sung et al.\cite{sung2015data} predefines several categories of semantic parts of 3D shapes, and uses geometric characteristics such as part positions, scales, and orientations to find similar parts for missing regions from shape database. Such kind of methods usually fails in the situation of more complicated shapes, which are beyond the description of the predefined semantic part categories or geometric characteristics. In contrast, deep learning based completion methods can learn more flexible features to predict a complete shape from an incomplete input. This kind of methods will be detailed in the subsection below.

\noindent\textbf{Deep learning approaches for 3D shape completion.}
The second category includes neural networks based methods, which take advantage of deep learning to learn the representation from the input shape \cite{han2019seqviews2seqlabels,wen2020point2spatialcapsule,han20193d2seqviews,liu2019sequence,Jiang2019SDFDiffDRcvpr} and predict the complete shape according to the representation, using an encoder-decoder framework. This category can be further classified according to different input shape forms including: volumetric shape completion \cite{dai2017shape,han2017high,stutz2018learning} and point cloud completion \cite{han2019multi,tchapmi2019topnet,sarmad2019rl,hu2019render4completion,wang2020cascaded,xin2021pmp}.
Our Cycle4Completion also falls into this category, which completes 3D shapes represented by point clouds.
Notable recent studies like CDN \cite{wang2020cascaded}, NSFA \cite{zhang2020detail} and SA-Net \cite{wen2020sa} have achieved impressive results on supervised point cloud completion task.
Moreover, RL-GAN-Net \cite{sarmad2019rl} introduced the reinforcement learning with the adversarial training to further improve the reality and consistency of the generated complete point clouds.
However, although great improvements have been made in supervised point cloud completion task, this task strongly depends on the paired training data, but the paired ground truth for incomplete real-world scan is rarely available. On the other hand, there is very few studies concerning the unpaired point cloud completion task.
As one of the pioneering work, AML \cite{stutz2018learning} directly measured the maximum likelihood between the latent representation of incomplete and complete shapes. Following the similar practice, Pcl2Pcl \cite{chen2019unpaired} introduced the GAN framework to bridge the semantic gap between incomplete and complete shapes. And Wu et al. \cite{wu2020multimodal} proposed a VAE based framework to predict multiple complete shapes for a single incomplete input.

Compared with the above-mentioned unpaired methods, our Cycle4Completion further establishes the self-supervision by cycle transformations in the latent space from both directions, which can provide a better guidance to learn the bidirectional geometric correspondence between incomplete shape and complete ones.

\noindent\textbf{Relationships with GANs.}
Our work is also related to the generative adversarial networks (GAN). Especially, our work is inspired by the unpaired style transferring network CycleGAN \cite{zhu2017unpaired} in 2D domain. However, it is usually difficult to directly apply a framework like CycleGAN to point cloud completion, where the simple cycle-consistency loss often fails to guide the generator to infer the missing shapes, because conceiving a consistent missing shape for the incomplete input is more complicated than transferring styles. Therefore, we propose to perform the cycle transformation in the latent space, where the partial and cycle matching losses are proposed for maintaining the transferred shapes consistency. Considering that 3D completion is essentially a reconstruction process from 3D shape to 3D shapes, the reconstruction of 3D shapes from 2D images \cite{Han2020TIP,Han2020Sdrwr,Han2020ECCV} is also a notable research direction, which is closely related to 3D completion. The difference between the two tasks is that 3D reconstruction from 2D images does not require 3D information as input, while the completion task based on 3D shapes requires 3D shape information as input.

\section{The Architecture of Cycle4Completion}


\subsection{Formulation}
\begin{figure*}[!t]
  \centering
  \includegraphics[width=\textwidth]{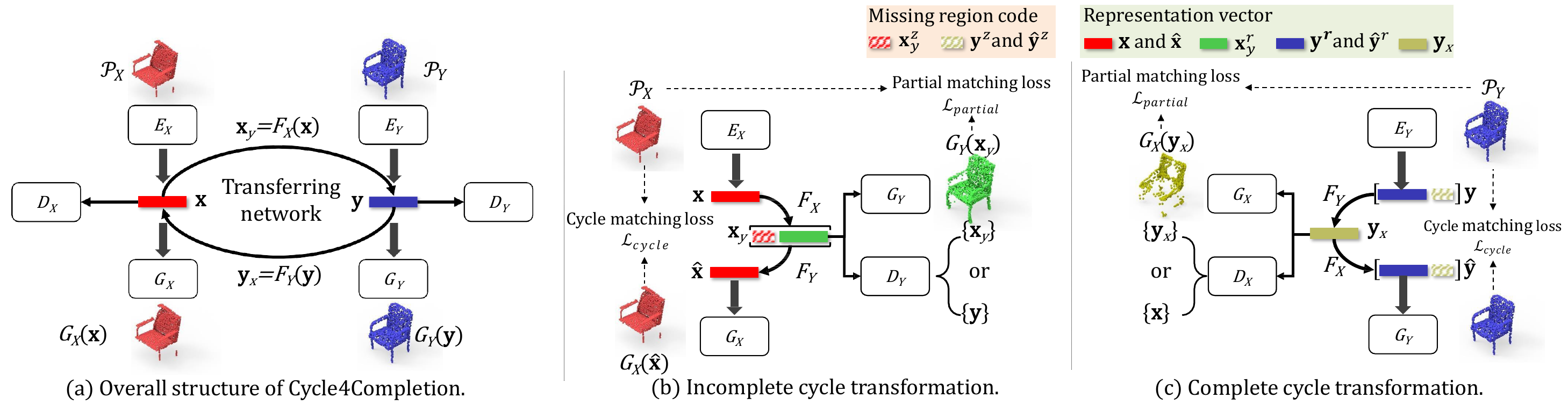}
  \caption{Illustration of Cycle4Completion. The overall structure in (a) consists of the incomplete cycle transformation in (b), which produces the complete prediction (green) from the incomplete input (red), and the complete cycle transformation in (c), which produces the incomplete prediction (yellow) from the complete input (blue). Both of the two cycles use self-reconstruction to learn the shape consistency.
  }
  \label{fig:overview}
\end{figure*}

We first describe the basic formulations in our method. As shown in Figure \ref{fig:overview}(a), let $\mathcal{P}_X\mbox{=}\{\mathbf{p}^x_i\}$ denote the point cloud of an incomplete shape, and $\mathcal{P}_Y\mbox{=}\{\mathbf{p}^y_i\}$ denote the point cloud of complete one. Our goal is to learn two mappings $F_X$ and $F_Y$ between the latent representations $\{\mathbf{x}\}$ of incomplete shapes and the latent representations $\{\mathbf{y}\}$ of complete shapes.
These representations are generated by the point cloud encoders $E_X: \mathcal{P}_X \rightarrow \mathbf{x}$ and $E_Y: \mathcal{P}_Y \rightarrow \mathbf{y}$, respectively, which are trained under the auto-encoder framework with the point cloud generators $G_X$ and $G_Y$, respectively.
In addition, two adversarial discriminators $D_X$ and $D_Y$ are introduced. $D_X$ aims to distinguish between $\{\mathbf{x}\}$ and $\{\mathbf{y}_x\}$, where $\mathbf{y}_x\mbox{=}F_Y(\mathbf{y})$. $D_Y$ aims to distinguish between $\{\mathbf{y}\}$ and $\{\mathbf{x}_y\}$, where $\mathbf{x}_y\mbox{=}F_X(\mathbf{x})$. We denote the compound operation of two functions $F_X$ and $F_Y$ as $F_XF_Y$.

\subsection{Encoder-decoder for Learning Latent Space}
The two auto-encoders learn the latent representation spaces for incomplete and complete shapes, respectively. We define the full Chamfer distance (CD) between two point clouds $\mathcal{P}_1$ and $\mathcal{P}_2$ as
\begin{equation}\small
  \mathcal{L}_{\rm CD}(\mathcal{P}_1\leftrightharpoons \mathcal{P}_2)\mbox{=}\sum_{\mathbf{p}^1_i\in \mathcal{P}_1}\min_{\mathbf{p}^2_i\in \mathcal{P}_2}\|\mathbf{p}^1_i-\mathbf{p}^2_i \|+\sum_{\mathbf{p}^2_i\in\mathcal{P}_2}\min_{\mathbf{p}^1_i\in\mathcal{P}_1}\|\mathbf{p}^2_i-\mathbf{p}^1_i \|.
\end{equation}
The reconstruction loss $\mathcal{L}_{AE}$ for training the auto-encoder framework is formulated as:
\begin{equation}\small
  \mathcal{L}_{AE}=\mathcal{L}_{CD}(\mathcal{P}_X\leftrightharpoons G_X(\mathbf{x}))+\mathcal{L}_{CD}(\mathcal{P}_Y \leftrightharpoons G_Y(\mathbf{y})).
\end{equation}

\subsection{Cycle Transformation}

\noindent\textbf{Transformation with missing region coding.}
For the incomplete cycle transformation in Figure \ref{fig:overview}(b), the missing region code $\mathbf{x}_y^z$ and the complete shape representation $\mathbf{x}_y^r$ are generated by $F_X$ when transferring $\mathbf{x}$ from the incomplete domain into the complete domain as $\mathbf{x}_y$. Therefore, $\mathbf{x}_y$ can be further denoted as $\mathbf{x}_y=[\mathbf{x}_y^r:\mathbf{x}_y^z]$. The remark ``:'' indicates the concatenation of two feature vectors. The complete shape is then predicted based on $\mathbf{x}_y^r$ by $G_Y$ as $G_Y(\mathbf{x}_y^r)$. And the discriminator $D_Y$ only discriminates between $\mathbf{x}_y^r$ and $\mathbf{y}$. In order to establish the shape consistency during the transformation process, $\mathbf{x}_y$ is projected back into the incomplete domain again by $F_Y$, denoted as $\mathbf{\hat{x}}$. The cycle reconstructed shape is predicted by $G_X$, denoted as $G_X(\mathbf{\hat{x}})$.

For the complete cycle transformation in Figure \ref{fig:overview}(c), the encoder $E_Y$ directly predicts a complete shape representation $\mathbf{y}^r$. In order to predict an incomplete shape, we randomly sample a missing region code from an uniform distribution between $[0,1]$, denoted as $\mathbf{y}^z$, and concatenate it with $\mathbf{y}^r$, denoted as $\mathbf{y}=[\mathbf{y}^r:\mathbf{y}^z]$. Then, the transformation network $F_Y$ transforms $\mathbf{y}$ into the incomplete domain, denoted as $\mathbf{y}_x$. Similar to the incomplete cycle transformation, the incomplete shape is predicted based on $\mathbf{y}_x$ by $G_X$, denoted as $G_X(\mathbf{y}_x)$. And the discriminator $D_X$ discriminates between $\mathbf{y}_x$ and $\mathbf{x}$. Following the inverse direction of incomplete cycle transformation, the shape consistency during the complete cycle transformation is established by predicting the reconstructed shape $G_Y(\mathbf{\hat{y}})$, where $\mathbf{\hat{y}}=F_X(\mathbf{y}_x)$. Note that same as $\mathbf{y}$, $\mathbf{\hat{y}}$ also consists of a complete representation $\mathbf{\hat{y}}^r$ and a missing region code $\mathbf{\hat{y}}^z$.

\noindent\textbf{Code matching Loss.}
In the complete cycle transformation in Figure \ref{fig:overview}(c), a missing region code $\mathbf{y}^z$ is sampled from a uniform distribution in order to create missing regions from the current complete input $\mathcal{P}_Y$. After the shape $\mathcal{P}_Y$ is cycled through $F_Y$ and $F_X$, a new missing region code $\mathbf{\hat{y}}_z$ is predicted by the transformation network $F_{Y}F_{X}$. Because both $\mathbf{y}^z$ and $\mathbf{\hat{y}}_z$ correspond to the same incomplete shape, the two codes should be equal. Therefore, we propose to use the Euclidean distance between $\mathbf{y}^z$ and $\mathbf{\hat{y}}_z$ as the \emph{code matching loss}, which can be formulated as:
\begin{equation}\small
    \mathcal{L}_{code}= \|\mathbf{y}^z - \mathbf{\hat{y}}_z\|^2.
\end{equation}

\noindent\textbf{Cycle matching loss.}
The \emph{cycle matching loss} aims to match the shapes of cycle reconstruction $G_Y(\hat{\mathbf{y}})/G_X(\hat{\mathbf{x}})$ to their corresponding input $\mathcal{P}_Y/\mathcal{P}_X$, which should keep the shape consistency throughout the whole transformation process.
Specifically, we define the cycle matching loss as the full Chamfer distance between the input $\mathcal{P}_Y/\mathcal{P}_X$ and the reconstructed point cloud $G_Y(\hat{\mathbf{y}})/G_X(\hat{\mathbf{x}})$ as $\mathcal{L}_{\rm CD}(\mathcal{P}_X \leftrightharpoons G_X(\hat{\mathbf{x}}))$ and $\mathcal{L}_{\rm CD}(\mathcal{P}_Y \leftrightharpoons G_Y(\hat{\mathbf{y}}))$, respectively.
Then we indicate the \emph{full cycle matching loss} for transferring network $F_X$ and $F_Y$ as:
\begin{equation}\small
  \mathcal{L}_{cycle}=\mathcal{L}_{\rm CD}(\mathcal{P}_X\leftrightharpoons G_X(\hat{\mathbf{x}})) + \mathcal{L}_{\rm CD}(\mathcal{P}_Y \leftrightharpoons G_Y(\hat{\mathbf{y}})).
\end{equation}


\noindent\textbf{Partial matching loss.}
The \emph{partial matching loss} is a directional constraint, which aims to match one shape to another without the matching in the inverse direction. Similar practice can be found in previous work \cite{chen2019unpaired}, which adopted the directional Hausdoff distance to partially match the complete prediction to the incomplete input.
However, the partial matching on the single direction cannot provide further guidance for the inference of missing regions, so we integrate the partial matching into the cycle transformation to establish a more comprehensive geometric correspondence on both directions.
We define the partial Chamfer distance between two point clouds $\mathcal{P}_1$ and $\mathcal{P}_2$ as:
 \begin{equation}\small
  \mathcal{L}_{\rm CD'}(\mathcal{P}_1\rightarrow \mathcal{P}_2)=\sum_{\mathbf{p}^1_i\in \mathcal{P}_1}\min_{\mathbf{p}^2_i\in \mathcal{P}_2}\|\mathbf{p}^1_i-\mathbf{p}^2_i \|.
\end{equation}
It is a constraint that only requires that the shape of $\mathcal{P}_2$ partially matches the shape of $\mathcal{P}_1$. For incomplete-cycle in Figure \ref{fig:overview}(b), the partial matching loss is formulated as $\mathcal{L}_{\rm CD'}(\mathcal{P}_X\rightarrow G_Y(\mathbf{x}^r_y))$, and for complete-cycle in Figure \ref{fig:overview}(c), the partial matching loss is formulated as $\mathcal{L}_{\rm CD'}(G_X(\mathbf{y}_x)\rightarrow \mathcal{P}_Y)$.
Note that the directions of above two partial Chamfer distances are always pointed from incomplete shapes to complete ones, which guarantees the incomplete shape partially matches the complete one, no matter whether it is predicted or real.
The \emph{full partial matching loss} is defined as:
\begin{equation}\small
  \mathcal{L}_{\rm partial}=\mathcal{L}_{\rm CD'}(\mathcal{P}_X\rightarrow G_Y(\mathbf{x}^r_y)) + \mathcal{L}_{\rm CD'}(G_X(\mathbf{y}_x)\rightarrow \mathcal{P}_Y).
\end{equation}




\noindent\textbf{Adversarial loss.}
To further bridge the geometric gap between the latent representations of complete and incomplete shapes, the adversarial learning framework is adopted as an unpaired constraint. Specifically, two discriminators $D_X$ and $D_Y$ are used to distinguish the real and fake representations in the incomplete and complete domains, respectively. The $D_X$ in incomplete domain discriminates between the real latent representations $\{\mathbf{x}\}$ and the fake latent representations $\{\mathbf{y}_x\}$; in the same way, the $D_Y$ in complete domain discriminates between $\{\mathbf{y}\}$ and $\{\mathbf{x}_y\}$.
In order to stabilize the training, we formulate the objective loss for discriminator under the WGAN-GP \cite{gulrajani2017improved} framework. For simplicity, we formulate the loss for $D_X$ as:
\begin{equation}\small
  \mathcal{L}_{D_X} = \mathbb{E}_{\mathbf{x}}D_X(\mathbf{x})-\mathbb{E}_{\mathbf{y}_x}D_X(\mathbf{y}_x)+\lambda_{gp}\mathcal{T}_{D_X},
\end{equation}
where $\lambda_{gp}$ is a pre-defined weight factor and $\mathcal{T}_{D_X}$ is gradient penalty term, denoted as:
\begin{equation}\small
  \mathcal{T}_{D_X}= \mathbb{E}_{\mathbf{x}}[({\| {\nabla}_{\mathbf{x}}D_X(\mathbf{x})\|}_2-1)^2].
\end{equation}
The discriminator loss $\mathcal{L}_{D_Y}$ for $D_Y$ can be formulated in the same way. The final adversarial losses for generator $\{F_X, F_Y\}$ and discriminator $\{D_X, D_Y\}$ are given as
\begin{equation}\small
  \mathcal{L}_D = \mathcal{L}_{D_X}+\mathcal{L}_{D_Y},
\end{equation}
\begin{equation}\small
  \mathcal{L}_G = \mathbb{E}_{\mathbf{y}_x}D_X(\mathbf{y}_x) + \mathbb{E}_{\mathbf{x}_y}D_Y(\mathbf{x}^r_y).
\end{equation}

\subsection{Training Strategy}
In our model, there are four sets of losses in total. We use $\Theta_D$ to denote the trainable parameters in $\{D_X,D_Y\}$, $\Theta_{AE}$ to denote the trainable parameters in $\{E_X, G_X, E_Y, G_Y\}$, and $\Theta_F$ to denote the trainable parameters in $\{F_X,F_Y\}$. We use $\mathcal{L}_G(\Theta_{AE}, \Theta_{F}, \Theta_{D})$ to denote that there are three parts of network (i.e. auto-encoder, transferring network, and discriminator) involved in calculating $\mathcal{L}_G$.

Given the learning rate $\gamma$, the encoder-decoder loss regularizes the parameters $\Theta_{AE}$, where the gradient optimization step is expressed as
\begin{equation}\small\label{eq:ed}
  \Theta_{AE} \leftarrow \Theta_{AE} -  \gamma\frac{\partial \mathcal{L}_{AE}(\Theta_{AE})}{\partial \Theta_{AE}}.
\end{equation}
The cycle matching loss and partial matching loss along with the adversarial loss regularize the transferring network. The gradient descent step for $\Theta_F$ is given as
\begin{equation}\small\label{eq:f}
\begin{split}
  \Theta_{F} \leftarrow &\Theta_{F} -  \gamma[\lambda_g\frac{\partial \mathcal{L}_G(\Theta_{AE}, \Theta_{F}, \Theta_{D})}{\partial \Theta_{F}} + \\
   &\lambda_p\frac{\partial \mathcal{L}_{\rm partial}(\Theta_{AE}, \Theta_{F})}{\partial \Theta_{F}} + \lambda_c\frac{\partial \mathcal{L}_{\rm cycle}(\Theta_{AE}, \Theta_{F})}{\partial \Theta_{F}}],
\end{split}
\end{equation}
where $\{\lambda_g, \lambda_c, \lambda_p\}$ are weight factors. Note that although $\mathcal{L}_G$ and $\mathcal{L}_{\rm partial}$ involve the parameter $\Theta_{AE}$, we fix the parameter $\Theta_{AE}$ when training $\mathcal{L}_G$ and $\mathcal{L}_{\rm partial}$.
The reason is that both $\mathcal{L}_G$ and $\mathcal{L}_{\rm partial}$ are constraints for the transformation process, while the two auto-encoders aim to learn a latent representation space instead of transferring features between complete and incomplete domains.
The weight factors are fixed to $\lambda_g$=1, $\lambda_c$=0.01 and $\lambda_p$=1 in our experiments.
Finally, the discriminators $D_X$ and $D_Y$ are regularized by the discriminator loss $\mathcal{L}_D$
\begin{equation}\small\label{eq:d}
  \Theta_{D} \leftarrow \Theta_{D} -  \gamma\frac{\partial \mathcal{L}_D(\Theta_{AE},\Theta_{F},\Theta_{D})}{\partial \Theta_{D}}.
\end{equation}
The pseudo code for training is given in \textbf{Algorithm} \ref{algo:alg}.

\begin{algorithm}[H]\small
  \caption{Pseudo code for training Cycle4Completion. The critic step $n_D$ is fixed to 3 during training.}
  \begin{algorithmic}[1]
    \While{model has not converged}
        \State Update $\Theta_{AE}$ following Eq.\ref{eq:ed}
      \For{$t = 0, ..., n_{D}$}
        \State Update $\Theta_{D}$ following Eq.\ref{eq:d}
      \EndFor
      \State Update $\Theta_{F}$ following Eq.\ref{eq:f}
    \EndWhile
  \end{algorithmic}
  \label{algo:alg}
\end{algorithm}

\begin{table*}[!t]\small
\centering
\caption{Point cloud completion comparison on ShapeNet dataset in terms of per point Chamfer distance $\times 10^{4}$ (lower is better).}
\begin{tabular}{l|c|c|cccccccc}
\toprule
Methods &Supervised &Average  &Plane    &Cabinet  &Car   &Chair   &Lamp   &Sofa    &Table    &Boat      \\ \midrule
3D-EPN \cite{dai2017shape} &Yes &29.1 &60.0  &27.0    &24.0    &16.0    &38.0    &45.0    &14.0    &9.0       \\
FoldingNet  \cite{yang2018foldingnet} &Yes &9.2  &2.4  &8.5    &7.2    &10.3    &14.1    &9.1    &13.6    &8.8      \\
PCN  \cite{yuan2018pcn} &Yes  &\textbf{7.6} &\textbf{2.0} &\textbf{8.0}    &\textbf{5.0}    &9.0    &13.0    &8.0    &10.0    &\textbf{6.0}       \\
TopNet  \cite{tchapmi2019topnet} &Yes &8.4  &2.5  &8.8    &5.9    &9.3    &12.0    &8.4    &13.5    &7.1      \\
SA-Net \cite{wen2020sa}  &Yes  &7.7  &2.2    &9.1    &5.6    &\textbf{8.9}    &\textbf{10.0}    &\textbf{7.8}    &\textbf{9.9}    &7.2 \\
\midrule
AE (baseline)\cite{chen2019unpaired} &No &25.4 &4.0 &37.0  &19.0  &31.0  &26.0   &30.0  &44.0  &12.0 \\
Pcl2Pcl \cite{chen2019unpaired} &No   &17.4  &4.0  &19.0    &10.0    &20.0    &23.0    &\textbf{26.0}    &26.0    &11.0    \\
Cycle4Completion (Ours)  &No  &\textbf{14.1}  &\textbf{3.1}  &\textbf{10.9}   &\textbf{7.5}    &\textbf{14.6}    &\textbf{16.7}    &26.7    &24.5    &9.1     \\
Cycle4Completion* (Ours)  &No  &14.3  &3.7  &12.6   &8.1    &\textbf{14.6}    &18.2    &26.2   &\textbf{22.5}    &\textbf{8.7}     \\
\bottomrule
\end{tabular}
\label{table:shapenet_complete}
\end{table*}

\section{Experiments}




\subsection{Evaluation on ShapeNet Dataset}
\noindent\textbf{Dataset.} Following the previous studies \cite{dai2017shape,chen2019unpaired}, we evaluate our methods on 3D-EPN dataset \cite{dai2017shape}, in order to fairly compare Cycle4Completion with the previous unpaired point cloud completion methods. For each 3D object, 8 partial point clouds are generated by back-projecting 2.5D depth images from 8 views into 3D. We uniformly sample only 2,048 points on the mesh surfaces for both the complete and partial shapes.

\begin{figure}[t]
  \centering
  \includegraphics[width=\columnwidth]{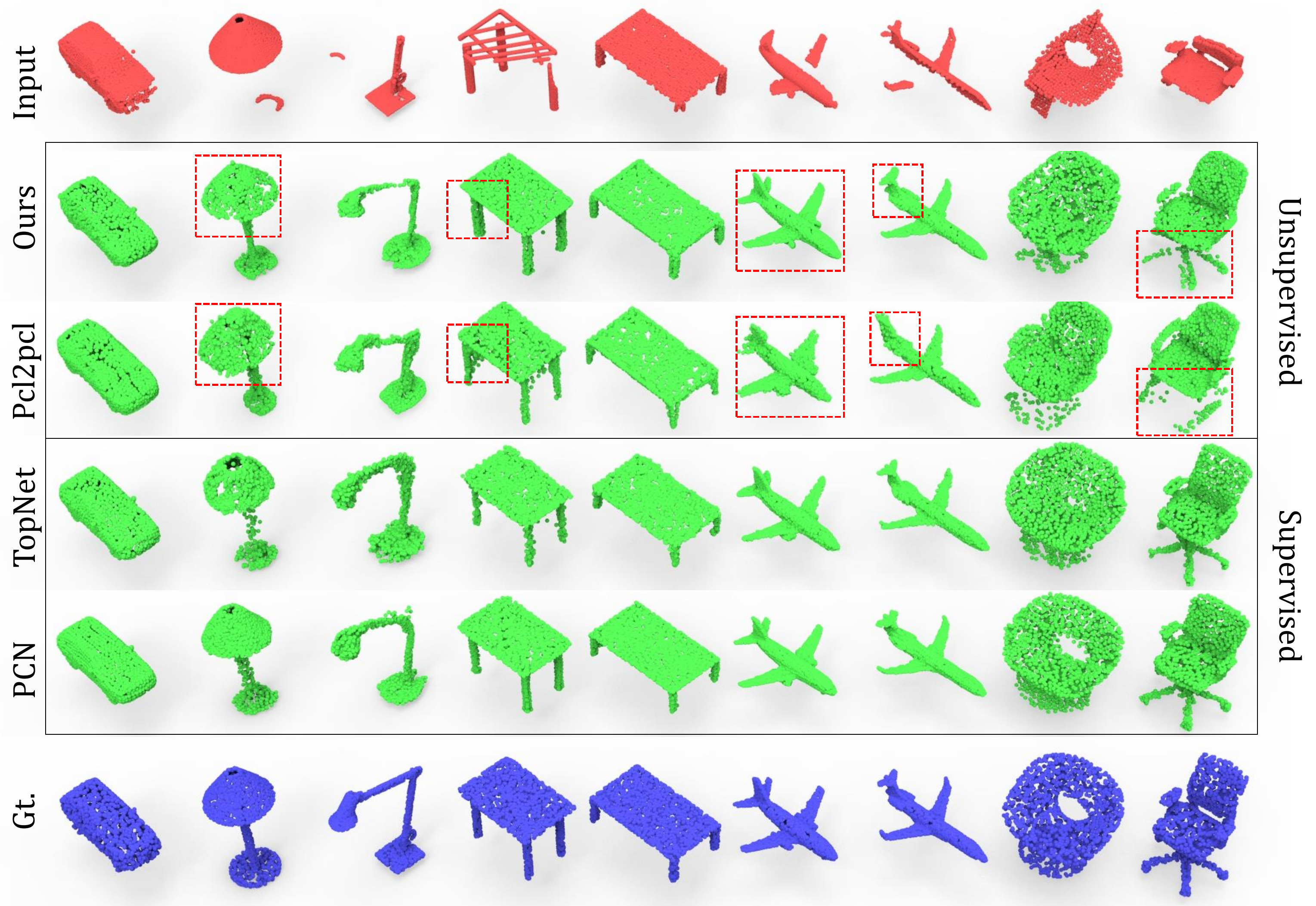}
  \caption{Visual comparison with the state-of-the-art completion methods on ShapeNet dataset.
  }
  \label{fig:shapenet_visual}
\end{figure}

\noindent\textbf{Quantitative and qualitative evaluation.} We use the per point Chamfer distance as the evaluation metric. In Table \ref{table:shapenet_complete}, we compare Cycle4Completion with some state-of-the-art supervised and unpaired point cloud completion methods. Since the training and testing split of Pcl2Pcl and 3D-EPN is different at ShapeNet dataset, we report our results on both of the two splittings for fair comparison. In Table \ref{table:shapenet_complete}, the \emph{Cycle4Completion} is the results of 3D-EPN splitting and the \emph{Cycle4Completion*} is the results of Pcl2Pcl splitting.
Moreover, we also quote the results of the baseline auto-encoder from \cite{chen2019unpaired} for comparison.
The experimental results show that our method achieves the best completion performance on all categories compared with the unpaired counterpart method Pcl2Pcl \cite{chen2019unpaired}.
And even comparing with the supervised methods, our Cycle4Completion still outperforms 3D-EPN \cite{dai2017shape} and yields a comparable results to PCN \cite{yuan2018pcn} and TopNet \cite{tchapmi2019topnet}.
In Figure \ref{fig:shapenet_visual}, we show the visualization results of point cloud completion using Cycle4Completion and compare it with other methods, from which we can find that our model predicts the complete shapes with higher accuracy than the unpaired Pcl2Pcl, especially on the regions highlighted by red rectangles. And the completion quality of our method is also comparable to the results of supervised methods.

\subsection{Evaluation on KITTI Dataset}
\begin{figure}[!t]
  \centering
  \includegraphics[width=\columnwidth]{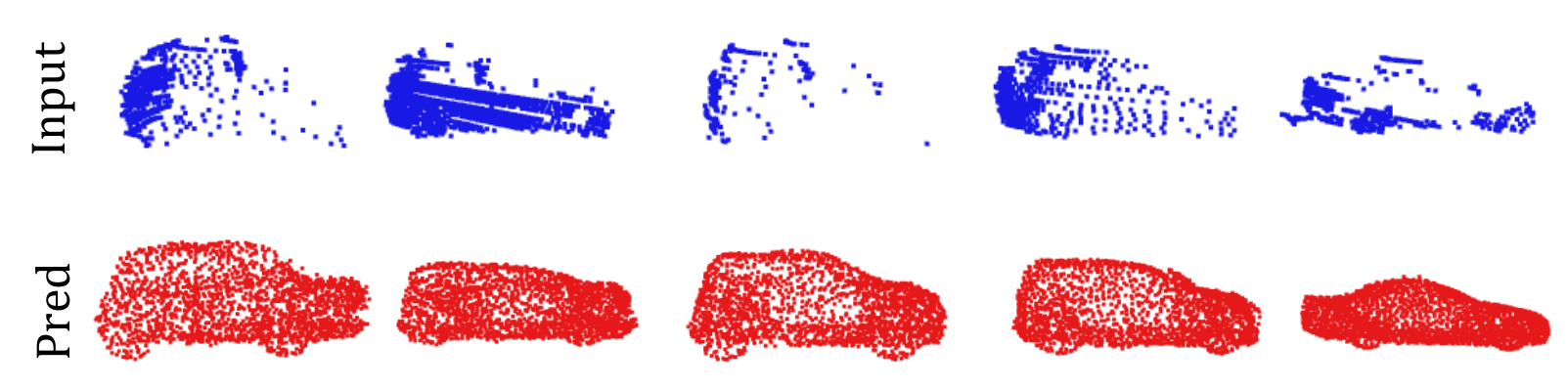}
  \caption{Visualizing the completion results of more similar car objects acquired from the KITTI dataset.
  }
  \label{fig:kitti_detail_rebutal}
\end{figure}

\begin{figure}[!t]
  \centering
  \includegraphics[width=\columnwidth]{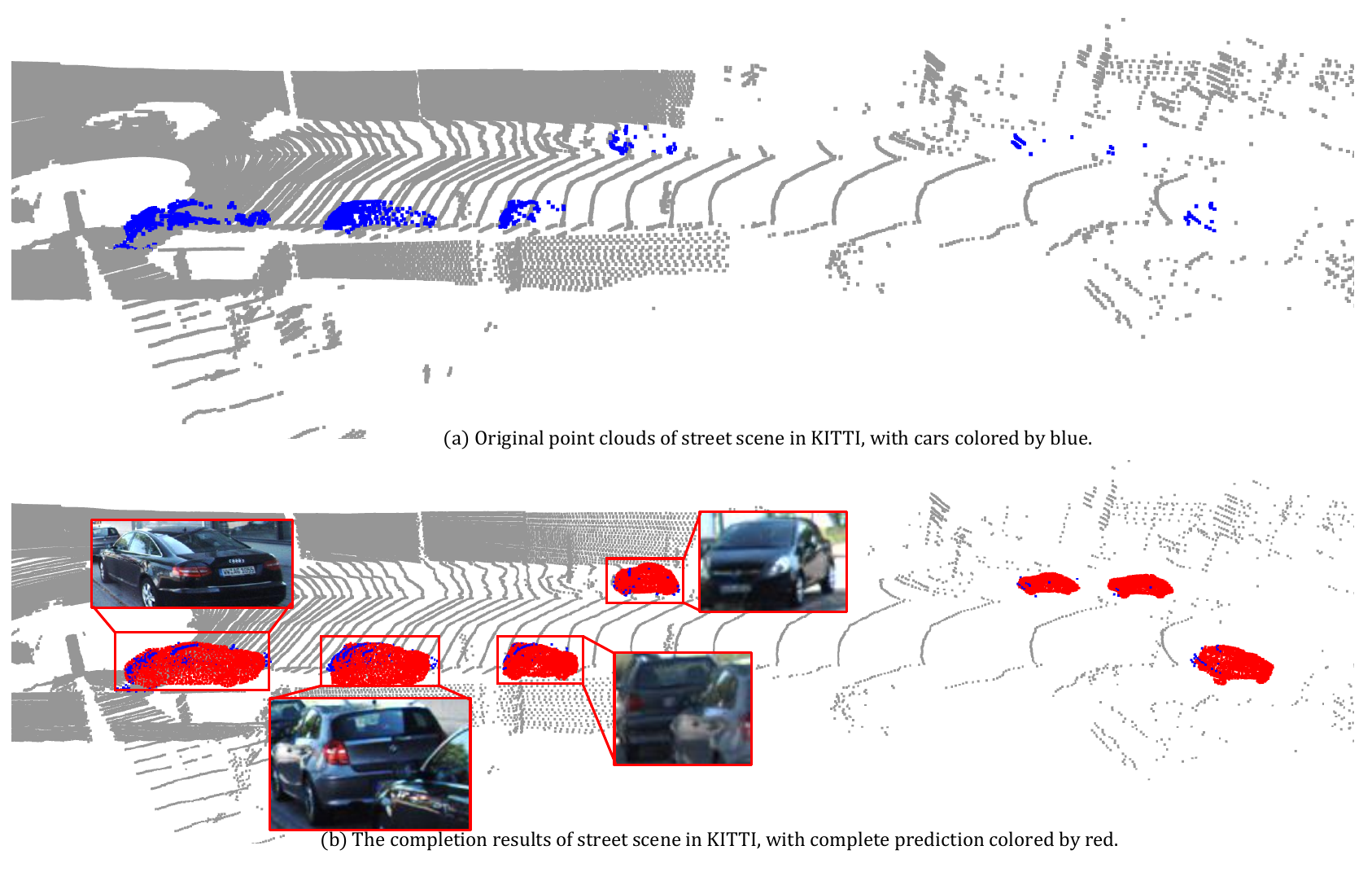}
  \caption{Visualizing the KITTI dataset with multiple completion cars.
  }
  \label{fig:kitti_scene_rebutal}
\end{figure}

We supplement the following qualitative results on the KITTI dataset, which contains car objects in real-world auto-navigation dataset scanned by LIDAR sensor in streets. The Cycle4Completion is first trained on ShapeNet dataset under car category, and then the trained Cycle4Completion is directly used to predict complete shapes on the KITTI dataset without any further fine-tuning process. In Figure \ref{fig:kitti_detail_rebutal}, we show the original incomplete point clouds of cars (highlighted with blue) directly obtained by LIDAR sensor, and the complete shape (highlighted with red) predicted by our Cycle4Completion. In Figure \ref{fig:kitti_scene_rebutal}, we further show the completion results integrated into the streets scene. Although there is no ground truth (i.e. the complete shapes of cars) for the KITTI dataset, we can still qualitatively find that Cycle4Completion predicts complete cars very robustly on the KITTI dataset, even our model is only trained on the ShapeNet dataset. This experiment shows that our Cycle4Completion model trained on ShapeNet dataset can achieve good completion results for more similar objects in the real-world scenario data.

\begin{figure}[!t]
  \centering
  \includegraphics[width=\columnwidth]{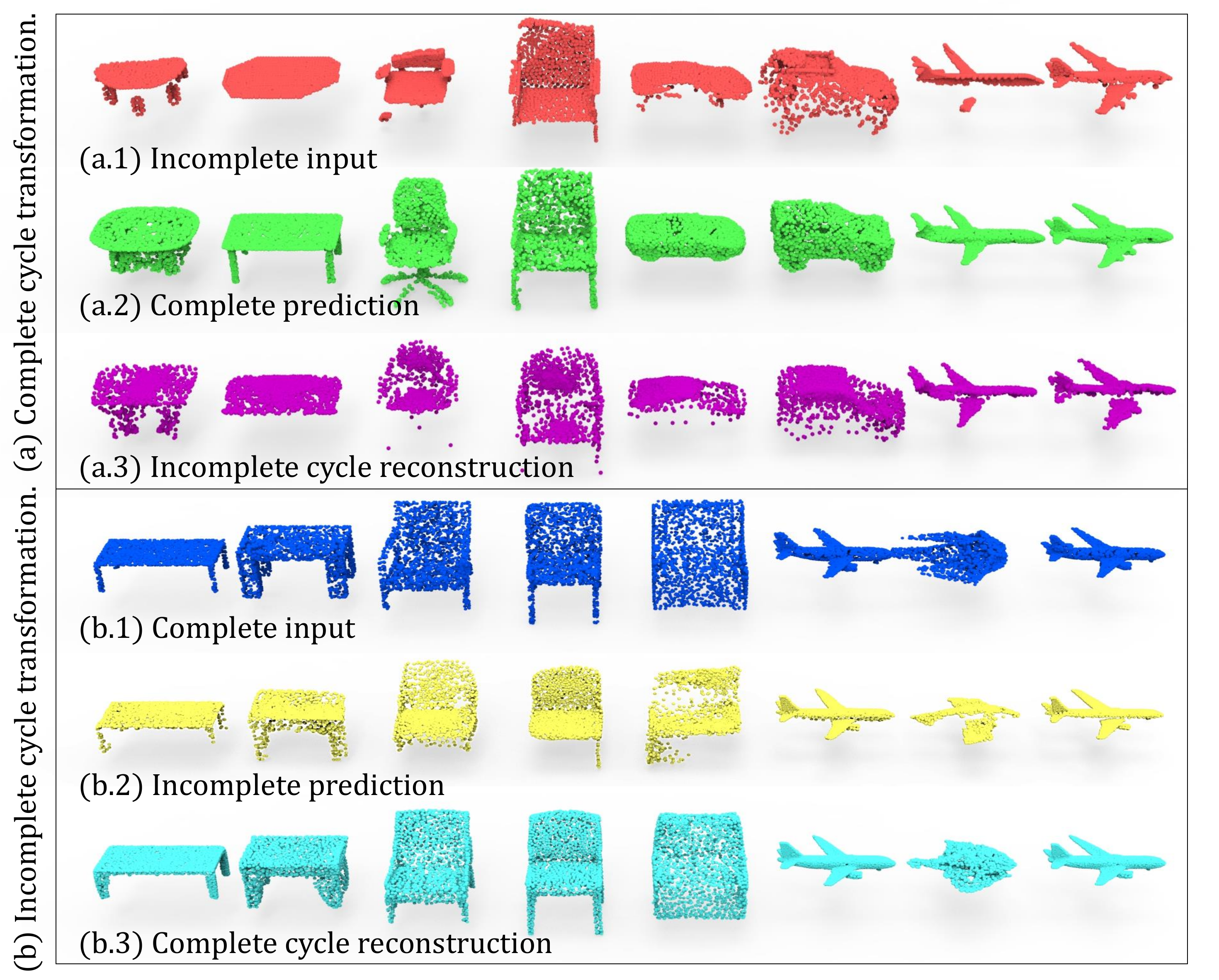}
  \caption{Visualization of the generated shapes in each step of cycle transformation.}
  \label{fig:cycle_vis}
\end{figure}

\subsection{Model Analysis}
For clarity, we typically analyze the performance of Cycle4Completion on four categories, i.e. plane, cabinet, car and chair.

\noindent\textbf{Visual analysis of complete cycle transformation.} We visualize the results of incomplete-cycle in Figure \ref{fig:cycle_vis}(a), and the results of complete-cycle in Figure \ref{fig:cycle_vis}(b). For incomplete-cycle, the input to the network is the incomplete shape, as shown in Figure \ref{fig:cycle_vis}(a.1). The complete prediction in Figure \ref{fig:cycle_vis}(a.2) demonstrates the effectiveness of Cycle4Completion to produce complete shapes from incomplete input, and the comparison between Figure \ref{fig:cycle_vis}(a.1) and Figure \ref{fig:cycle_vis}(a.3) proves that Cycle4Completion successfully learns to keep shape consistency throughout the whole cycle transformation. Similar conclusions can also be drawn from Figure \ref{fig:cycle_vis}(b) for the complete-cycle.

\begin{figure}[h]
  \centering
  \includegraphics[width=\columnwidth]{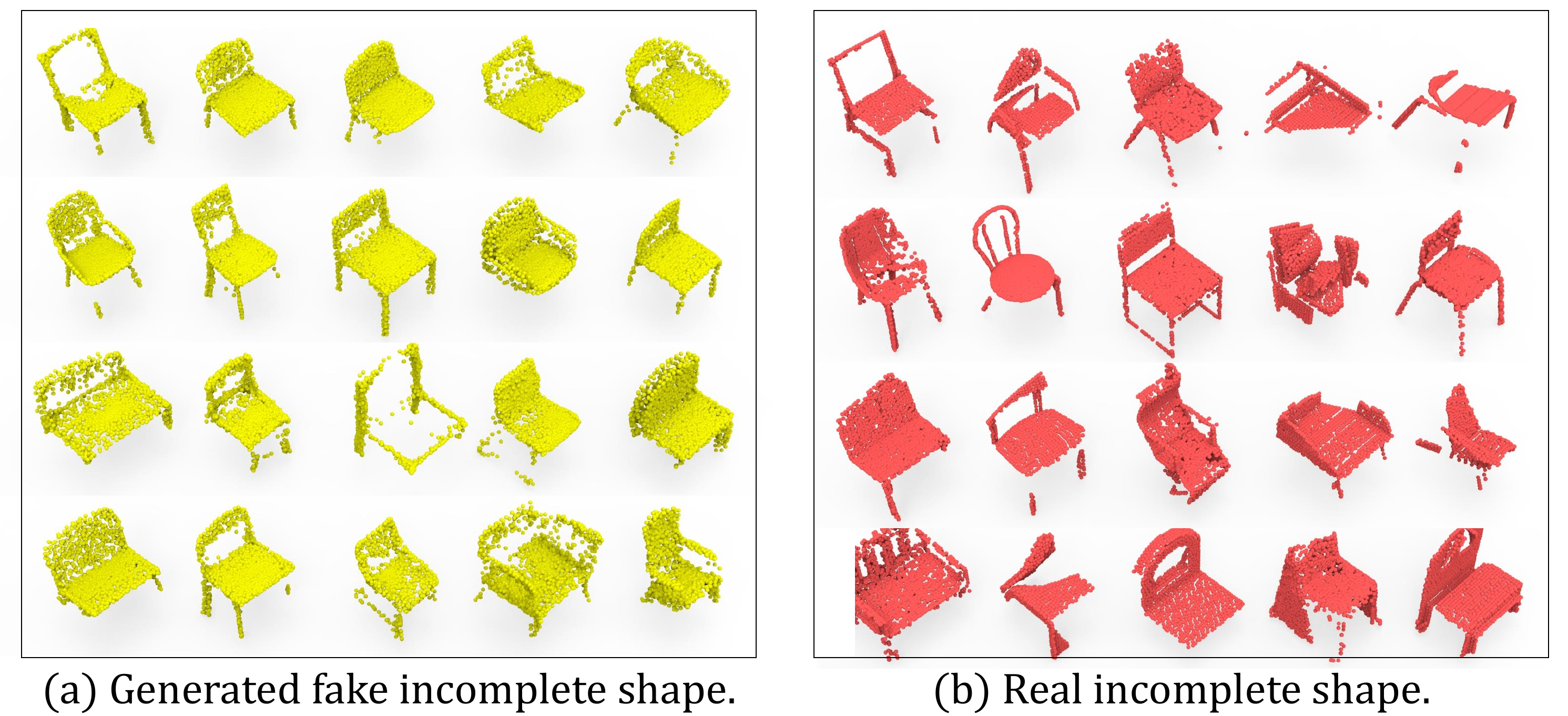}
  \caption{Visualization comparison of predicted incomplete shapes in (a) with the real incomplete shapes in (b).
  }
  \label{fig:partial_vis}
\end{figure}

\noindent\textbf{Visual analysis of incompletion quality.}  In Figure \ref{fig:partial_vis}, we visually evaluate the quality of incomplete shapes, which are generated by Cycle4Completion from the complete input on a specific category of chair. The visual comparison between the incomplete prediction in Figure \ref{fig:partial_vis}(a) and the real incomplete shapes in Figure \ref{fig:partial_vis}(b) proves that Cycle4Completion successfully learns the geometric correspondence from the complete shapes to the incomplete ones. The similar pattern of incompleteness to the real incomplete shapes justifies the good 3D shape understanding ability of our model.

\begin{figure}[h]
  \centering
  \includegraphics[width=\columnwidth]{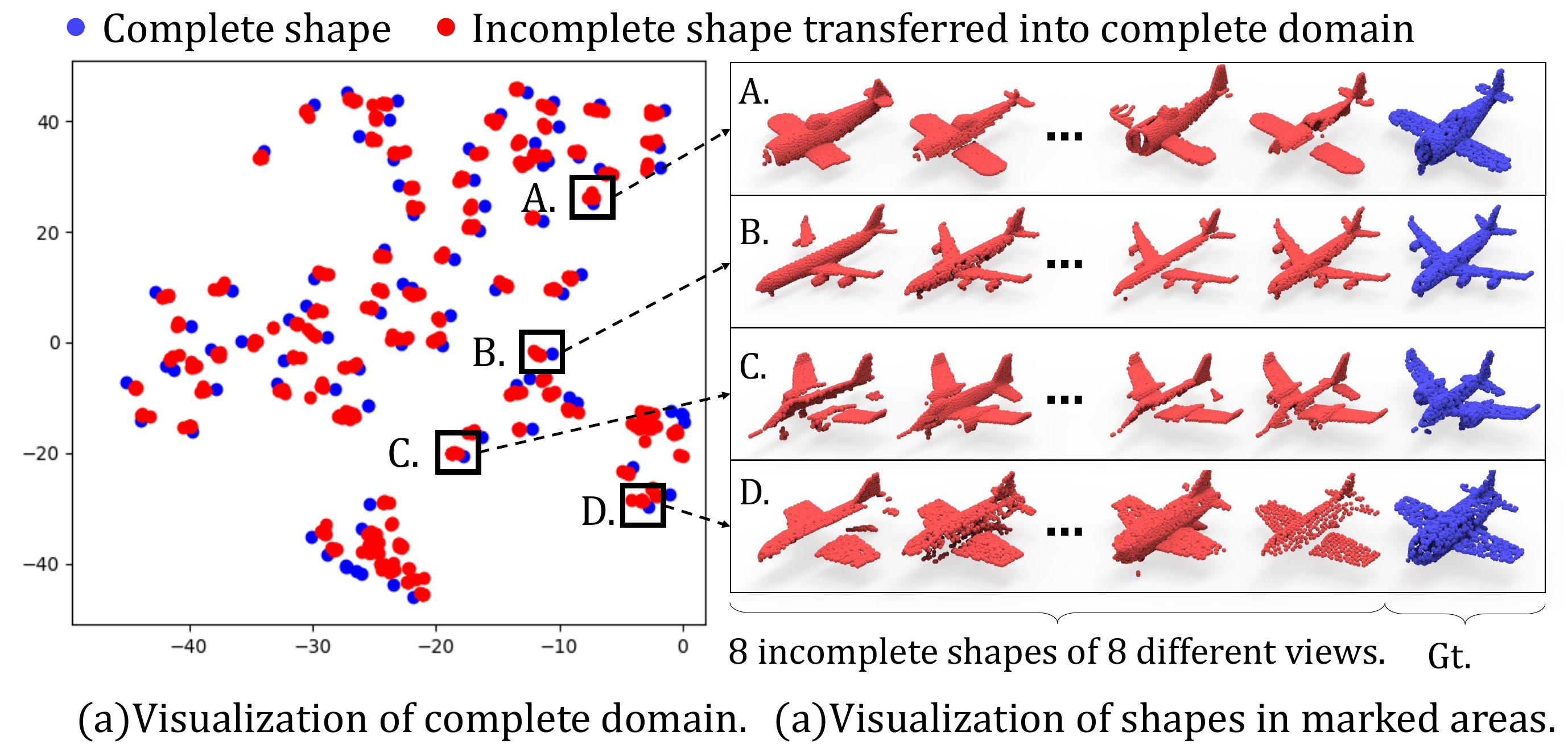}
  \caption{Visualization of the latent representation space in complete domain. We randomly choose four areas in (a) and visualize the shape represented by these points in (b).
  }
  \label{fig:latent_vis}
\end{figure}

\noindent\textbf{Visual analysis of latent space.} In Figure \ref{fig:latent_vis}(a), we use t-SNE\cite{maaten2008visualizing} to visualize the latent features of complete shape and the ones of incomplete shapes that are transferred from incomplete domain into the complete domain. The red points stand for the latent features of incomplete shapes, and the blue points stand for the complete ones. Note that the incomplete shape is generated from partial views of complete ones, and we have 8 partial shapes generated from 8 different views of each complete shape.
In Figure \ref{fig:latent_vis}(a), we can find that red and blue points are arranged in a \emph{paired pattern}, and from Figure \ref{fig:latent_vis}(a) we can find that in each local area highlighted by black rectangles, there is a pair of one complete shape and its 8 partial incomplete shapes, which are exactly the geometric corresponding pattern between complete and incomplete shapes in the dataset.
The visualization of latent space shows the effectiveness of Cycle4Completion to establish a well arranged latent space, and the ability to capture the shape correspondence between the complete and incomplete ones.

\begin{figure}[!t]
  \centering
  \includegraphics[width=\columnwidth]{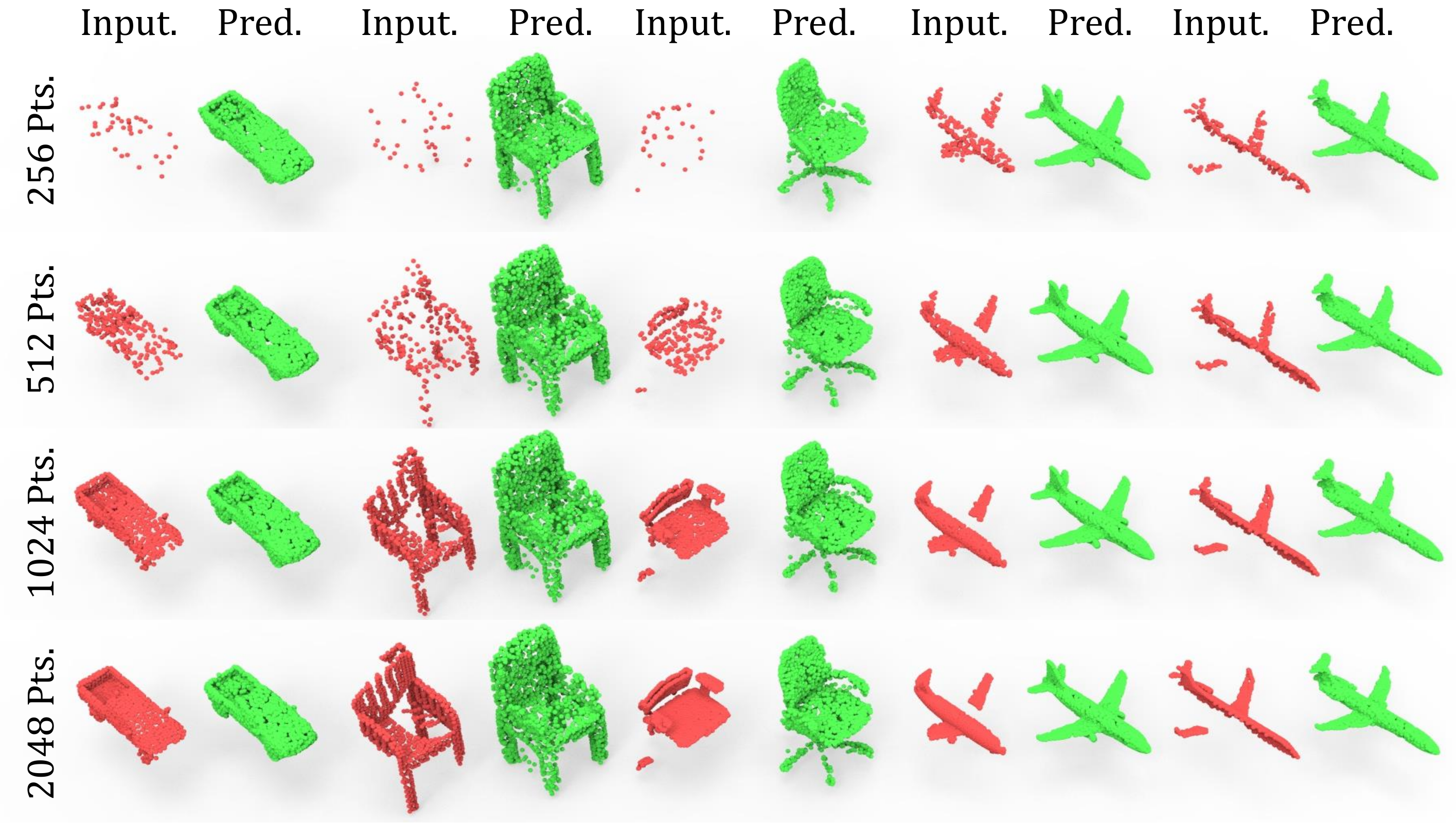}
  \caption{Visualization of completion performance under different input point numbers.
  }
  \label{fig:pt_num_vis}
\end{figure}

\noindent\textbf{Effect of each module to our model.} In order to analyze the effect of each module to our model, we develop three variations by removing element from Cycle4Completion including: (1) \emph{w/o Partial} is the variation that removes the partial matching loss; (2) \emph{w/o GAN} is the variation that removes the discriminator and its corresponding adversarial loss; (3) \emph{w/o Cycle} is the variation that removes the cycle matching loss; (4) \emph{w/o Coding} is the variation that removes all the missing region codes and the corresponding code matching loss. The results are shown in Table \ref{table:ablation_study}, where the \emph{Full model} represents the original version of Cycle4Completion. In Table \ref{table:ablation_study}, the Full Model achieves the best completion performance, which proves the contributions of each part to the performance of Cycle4Completion. Moreover, \emph{w/o Partial} variation yields the worst completion performance. This is because the model loses its supervision for learning the shape consistency when transferring representations from incomplete domain to complete one.
\begin{table}[h]\small
\centering
\caption{The effect of each part (per point CD $\times 10^{4}$).}
\begin{tabular}{l|c|cccc}
\toprule
Methods  &Average  &Plane    &Cabinet  &Car   &Chair       \\ \midrule
w/o Partial  &23.7 &15.6  &27.8    &14.8    &36.6        \\
w/o GAN   &12.8 &4.7 &18.4    &9.1    &19.0         \\
w/o Cycle  &10.4  &3.6  &12.5    &8.9    &15.8        \\
w/o Coding  &9.4  &3.2  &11.8   &7.7     &14.8     \\ \midrule
Full Model  &\textbf{9.1}  &\textbf{3.1}  &\textbf{10.9}   &\textbf{7.5}    &\textbf{14.6}     \\
\bottomrule
\end{tabular}
\label{table:ablation_study}
\end{table}

\noindent\textbf{Effect of input point number.}
In order to further evaluate the performance of Cycle4Completion on more sparse input, we evaluate Cycle4Completion using the input of partial point clouds with different resolutions. Specifically, we keep the number of 2048 points on the output complete shape unchanged, and evaluate the performance of Cycle4Completion on the input point clouds with resolutions ranging from 256 to 2048. The quantitative completion results are given in Table \ref{table:pt_num}, and the visualization results are shown in Figure \ref{fig:pt_num_vis}. Both quantitative and qualitative results demonstrate a robust performance of Cycle4Completion on various input resolutions.

\begin{table}[h]\small
  \centering
  \caption{The effect of input point number (CD $\times 10^{4}$).}
    \begin{tabular}{c|c|cccc}
    \toprule
    \#Points  &Average  &Plane    &Cabinet  &Car   &Chair       \\ \midrule
    256  &14.4 &3.3 &15.9    &9.9    &28.4        \\
    512   &10.0 &3.2 &12.1    &7.9    &16.9        \\
    1024  &9.6  &\textbf{3.2}  &11.9    &7.9    &15.5        \\ \midrule
    2048  &\textbf{9.1}  &\textbf{3.1}  &\textbf{10.9}   &\textbf{7.5}    &\textbf{14.6}     \\
    \bottomrule
    \end{tabular}
  \label{table:pt_num}
\end{table}

\noindent\textbf{Effect of different training strategies.} In our model, the generator loss $\mathcal{L}_G$, cycle matching loss $\mathcal{L}_{cycle}$ and partial matching loss $\mathcal{L}_{partial}$ involves multiple parameter sets. We selectively update some of the parameter sets while remain the others unchanged when training these losses. To evaluate the effectiveness of other potential training strategies, we develop the variation of (a) $\partial \mathcal{L}_G/\partial(\Theta_{AE},\Theta_{F})$ which updates both $\Theta_{AE}$ and $\Theta_{F}$ when training generator loss. Similarly, we also develop the variations of (b) $\partial \mathcal{L}_{partial}/\partial(\Theta_{AE},\Theta_{F})$ and (c) $\partial \mathcal{L}_{cycle}/\partial(\Theta_{AE},\Theta_{F})$ and report the results in Table \ref{table:training}. We observe a severe mode collapse in both variations (a) and (b), which is caused by training the auto-encoder cross different domain (both $\mathcal{L}_{partial}$ and $\mathcal{L}_{AE}$ involve transformation from one domain to the other). In contrast, $\mathcal{L}_{cycle}$ is a regularization considering one domain, and only involves reconstruction to input.

\begin{table}[h]\small
  \centering
  \caption{The effect of training strategies (CD $\times 10^{4}$)..}
  \resizebox{\linewidth}{!}{
  \begin{tabular}{l|c|cccc}
    \toprule
    Strategies &Average  &Plane    &Cabinet  &Car   &Chair       \\ \midrule
    $\partial \mathcal{L}_G/\partial(\Theta_{AE},\Theta_{F})$ &collapsed &-  &-    &-    &-        \\
    $\partial \mathcal{L}_{partial}/\partial(\Theta_{AE},\Theta_{F})$  &collapsed &-  &-    &-    &-   \\
    $\partial \mathcal{L}_{cycle}/\partial(\Theta_{AE},\Theta_{F})$  &10.1 &3.3  &12.5    &9.4    &15.1        \\\midrule
    Original   &\textbf{9.1}  &\textbf{3.1}  &\textbf{10.9}   &\textbf{7.5}    &\textbf{14.6}      \\
    \bottomrule
  \end{tabular}}
  \label{table:training}
\end{table}

\begin{figure}[t]
  \centering
  \includegraphics[width=0.9\columnwidth]{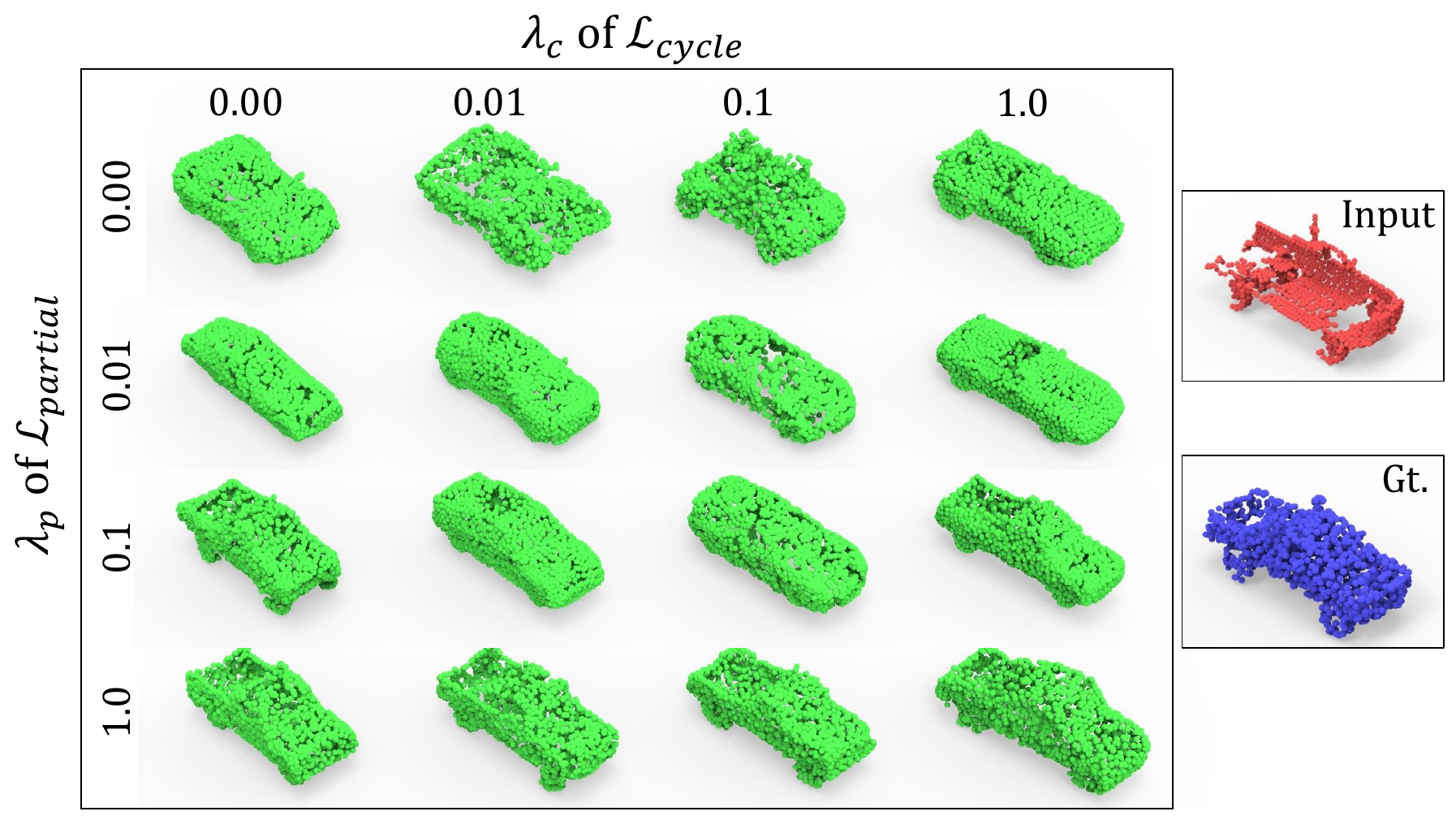}
  \caption{Visualization comparison of completion performance under different weight parameters $\lambda_{p}$ and $\lambda_{c}$.
  }
  \label{fig:ratio_comparison}
\end{figure}
\noindent\textbf{Effect of $\lambda_c$ and $\lambda_p$.} 
 The cycle matching loss tends to keep shape consistent throughout the whole cycle transformation, while the partial matching loss only keeps a single side consistency. Different ratio between $\lambda_c$ and $\lambda_p$ will result in different preference of model to establish the shape consistency. In Table \ref{table:ratio_comparison}, we quantitatively analyze the effect of weight factors $\lambda_c$ and $\lambda_p$ to our model on the specific car class, and in Figure \ref{fig:ratio_comparison}, we visually evaluate the corresponding completion performance. Since the completion task is a single-side transformation, a larger weight of $\lambda_p$ yields better performance. However, as shown in Table \ref{table:ablation_study}, totally removing partial matching loss ($\lambda_p$=0) will degrade the performance of our model.

\begin{table}[h]\small
  \centering
  \caption{The effect of $\lambda_{p}$ and $\lambda_{c}$ (CD $\times 10^{4}$).}
    \begin{tabular}{l|cccc}
    \toprule
    \diagbox{$\lambda_{p}$}{$\lambda_{c}$}  &0.0   &$10^{\mbox{-}2}$  &$10^{\mbox{-}1}$   &1.0       \\ \midrule
    0.0  &18.8  &17.1   & 16.9   &14.8        \\
    $10^{\mbox{-}2}$ &15.8  &15.6    &12.4    &12.4        \\
    $10^{\mbox{-}1}$  &9.8  &10.2    &9.8    &9.4   \\
    1.0  &8.9  &\textbf{7.5}    &8.1    &9.3        \\
    \bottomrule
    \end{tabular}
  \label{table:ratio_comparison}
\end{table}

\section{Conclusions}
We propose the Cycle4Completion for unpaired point cloud completion task.
Our model successfully captures the bidirectional geometric correspondence between incomplete and complete shapes, which enables the learning of point cloud completion without the paired complete shapes. Our model effectively learns to generate fake incomplete shapes to guide the completion network.
The proposed Cycle4Completion is evaluated on the widely used ShapeNet dataset, and the experimental results demonstrate the state-of-the-art performance compared with other unpaired completion methods.

{\small
\bibliographystyle{ieee_fullname}
\bibliography{ref}
}

\end{document}